%% file: root.tex
\documentclass[letterpaper, 10 pt, conference]{ieeeconf}  

\IEEEoverridecommandlockouts                              

\usepackage{subfigure} 
\usepackage{tikz}
\usepackage{pgfplots}
\usepackage{graphicx} 
\usepackage{amsmath} 

\title{\LARGE \bf
Quantifying Performance of Bipedal Standing with Multi-channel EMG
}

\author{Yanan Sui, Kun ho Kim, Joel W. Burdick
\thanks{Yanan Sui, Kun ho Kim and Joel W. Burdick are with California Institute of Technology, Pasadena, CA, USA.
        {\tt\small ysui@caltech.edu, khkim@caltech.edu, jwb@robotics.caltech.edu}. This version is published in IROS'17.}%
}

\begin{document}

\maketitle
\thispagestyle{empty}
\pagestyle{empty}

\begin{abstract}
Spinal cord stimulation
has enabled humans with motor complete spinal cord injury (SCI) to independently stand and recover some lost autonomic function.  Quantifying the quality of bipedal standing under spinal stimulation is important for spinal rehabilitation therapies and for new strategies that seek to combine spinal stimulation and rehabilitative robots (such as exoskeletons) in real time feedback.  To study the potential for automated electromyography (EMG) analysis in SCI, we evaluated the standing quality of paralyzed patients undergoing electrical spinal cord stimulation using both video and multi-channel surface EMG recordings during spinal stimulation therapy sessions. The quality of standing under different stimulation settings was quantified manually by experienced clinicians. By correlating features of the recorded EMG activity with the expert evaluations, we show that multi-channel EMG recording can provide accurate, fast, and robust estimation for the quality of bipedal standing in spinally stimulated SCI patients. Moreover, our analysis shows that the total number of EMG channels needed to effectively predict standing quality can be reduced while maintaining high estimation accuracy, which provides more flexibility for rehabilitation robotic systems to incorporate EMG recordings. 
\end{abstract}

\input{sections/introduction}

\input{sections/related}

\input{sections/methods}

\input{sections/results}

\input{sections/conclusions}




\section*{ACKNOWLEDGMENT}

This work was supported by the Helmsley Foundation, the Christopher and Dana Reeve Foundation, and the National Institutes of Health (NIH). The authors thank E. Rejc, C. Angeli and S. Harkema for help with the experiments.

\bibliographystyle{abbrv}
\bibliography{bipedal_standing}

\end{document}

%% file: sections/introduction.tex
\section{INTRODUCTION}\label{sec:introduction}

\begin{figure}[b!]
\vskip -0.2 true in
\centering
\includegraphics[width=0.4\textwidth]{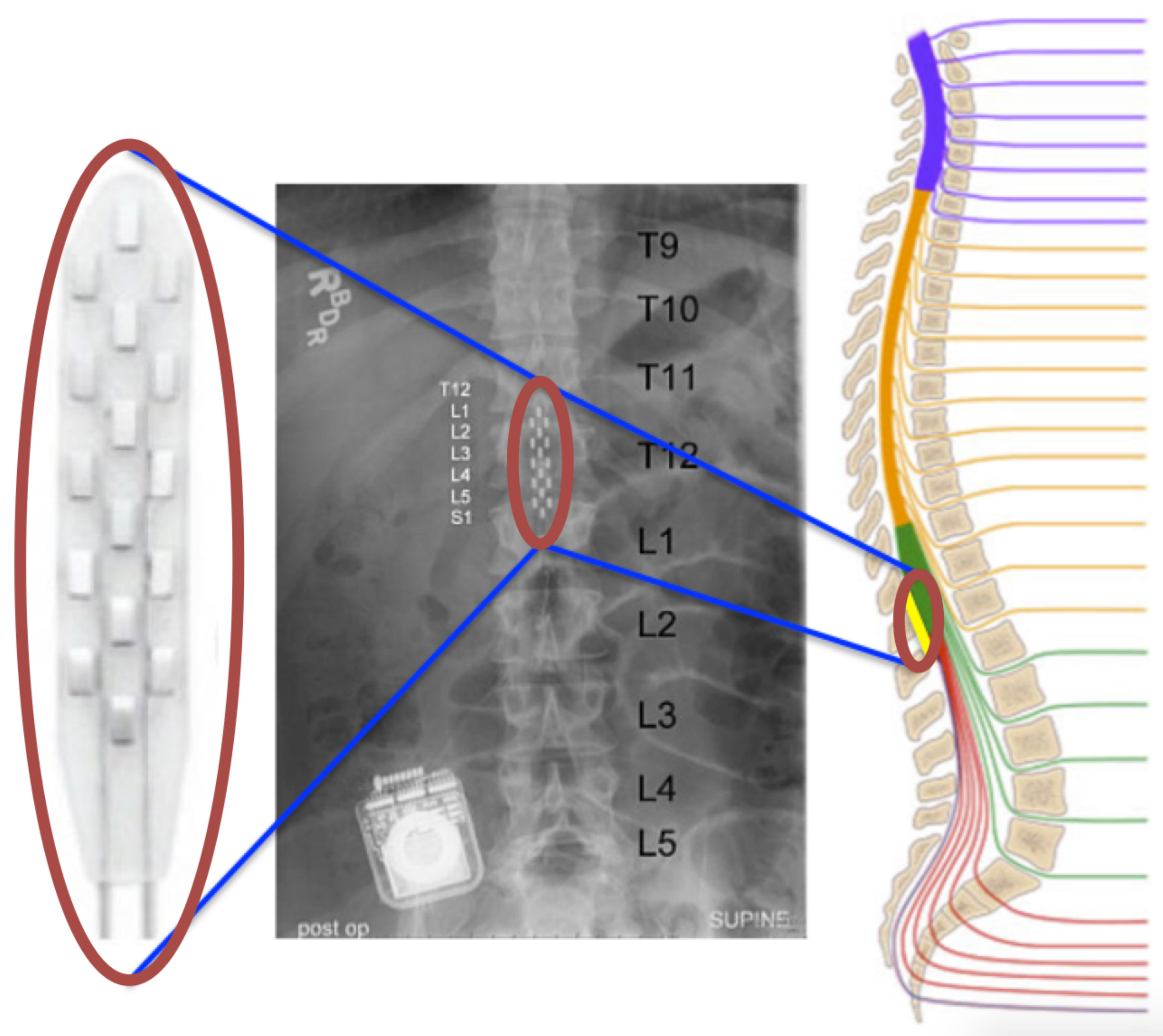}
\vskip -0.1 true in
\caption{Spinal Cord Stimulation with 16-electrode array implantation.}
\label{fig:array}
\end{figure}

Spinal Cord Injury (SCI) is a debilitating condition that afflicts $\sim$350,000 people in the U.S., and 5 million worldwide.  Electrical spinal stimulation, using multi-electrode arrays implanted in the epidural space over the lumbosacral spinal cord (see Fig. \ref{fig:array}), has enabled motor complete paralyzed SCI sufferers to achieve independent weight bearing standing, some weight-assisted stepping, and partial recovery of lost autonomic function.  These preliminary studies have shown that proper physical therapy must be combined with spinal stimulation to achieve the best  recovery, and that the stimulating parameters (which combinations of electrodes are activated, as well as their stimulating voltage or current amplitude, and stimulating frequency) which provide the best motor performance can vary substantially across patients.  Surface electromyographic (EMG) recordings obtained during training and therapy sessions play a valuable role in assessing the rate of patient progress under spinal stimulation, as well as the current ad hoc process of optimizing the electrical stimulation therapy for each patient.

This paper presents the first study of surface EMG signals in spinally stimulated SCI patients.  An  understanding of their properties will benefit  ongoing and future efforts in SCI rehabilitation.  We have shown how EMG signals can be used with machine learning algorithms to automatically optimize multi-electrode array stimulation parameters \cite{desautels2015active} in animal models of SCI.   Real-time EMG-based quantification of electrically stimulated standing performance could also provide feedback for the control of rehabilitative robotic devices, such as the Lokomat trainer or exoskeletons, which are coupled with spinal stimulation.  Recent experiments \cite{gerasimenko2015transcutaneous} show that the combination leads to synergistic outcomes.

The muscle activity in \textit{spinally stimulated standing} (SSS) need not be similar to that of healthy human subjects during quiet standing.  Compared to healthy standing, SSS has several distinguishing characteristics: \\
(1) SSS is mainly controlled by the activity of the implanted stimulating electrode array, rather than via the patient's voluntary motor control system .\\
(2) The activity of major muscle groups can be substantially different from the activity under natural healthy standing.\\
(3) Balance is more difficult to achieve in SSS.\\
(4) Common physical measures like center of mass, center of pressure are very dynamic compared to normal standing.\\

Our human subjects are implanted with a 16-electrode array (5-6-5 Specify, Medtronics) in the epidural space over the lumbosacral spinal cord (Fig. \ref{fig:array}). The electrical stimulation process consists of a sequence of electrical pulse trains applied to a set of selected electrodes. The possible stimulus patterns (selected electrodes, their polarity, the pulse amplitude and width, and the frequency of pulse train) generate a huge space of parameters we can choose from. 12-channel EMG were recorded from key postural muscles while subjects underwent various stimulation patterns.

This work is part of the preliminary study of helping paralyzed patients to stand via spinal cord stimulation and rehabilitation robotics. In this study, we tested the effectiveness of spinal cord stimulation on a clinically sensory and motor complete participant. It showed that the patient was able to stand over-ground bearing full body-weight without external assistance, only using hands to assist balance. We tested a large number of different stimuli. The qualities of standing are quite different under different stimulating parameters. Currently, the theory on how the muscle activities are  supposed to change under spinal stimulation is largely unknown. To our knowledge, this paper is the first attempt to quantify standing performance of SCI patients under epidural stimulation using multi-channel surface EMG.  We show that even with a limited number of features and simple linear prediction models, a 12-channel EMG recording can provide accurate, fast and robust estimation for the quality of bipedal standing. We also demonstrate with a Support Vector Machine (SVM) that using a more elaborate feature set can provide fine resolution predictions of standing performance. Moreover, we show through computational analysis that the total number of EMG channels can be significantly reduced while keeping a high accuracy for standing quality estimation. It helps to better understand the motor control mechanisms of SCI patients with spinal cord stimulation. This also suggests the possibility of using EMG to predict standing quality for real-time control of robotic prostheses.

%% file: sections/related.tex
\section{RELATED WORK} \label{sec:related}

Electrical stimulation can be used in many ways to enable or improve the motor function of SCI patients.  The data analyzed in this paper is relevant to the process of epidural spinal stimulation for human standing recovery.  It has been shown  \cite{harkema2011effect} \cite{rejc2015effects} that properly applied stimulation can enable paralyzed patients to achieve full weight-bearing standing.  The results obtained in this intervention \textit{are not} derived by direct stimulation of specific postural muscles, but by excitation of natural postural control circuits.

Functional Electrical Stimulation (FES), where electrical currents are applied to the peripheral motor nerves of paralyzed muscles to elicit muscle contractions, can provide significant levels of  motor function \cite{peckham2005functional}. It is widely used after SCI to enhance muscle strength and movements. EMG signals are used for on-line control of FES \cite{frigo2000emg}. Posture shifting after SCI using functional neuro-muscular stimulation has been studied in computer simulation \cite{audu2011posture}. Unlike FES, which has a direct mapping between neuro-muscular stimulation and muscle activity, the mapping between spinal cord stimulation and muscle activity is incompletely unknown.   However, EMG activity is important to the use of both of these electrical stimulation modalities in SCI.

Methods such as time-domain and frequency-domain analyses have been widely utilized in EMG pattern recognition \cite{phinyomark2012feature}.
Using EMG for movement intent prediction and control of robotic prostheses has been widely studied, for example, learning control of a robotic hand \cite{bitzer2006learning} or a wrist exoskeleton \cite{khokhar2010surface}.
EMG signals have also been used to enable paralyzed patients to control rehabilitation exoskeletons \cite{yin2012emg}, but not under the condition of spinal stimulation.

Biomechanical models are often used to simulate human standing and movement. They range from simple inverse pendulum models \cite{winter2009biomechanics}, to more complicated musculo-skeletal models such as \cite{geyer2010muscle}, \cite{wang2012optimizing}, \cite{mordatch2013animating}. Biomechanics and motor control of human movement are studied to better understand biological mechanisms, develop humanoid robots, and produce persuasive virtual animations of human beings. The motor control mechanisms of severe SCI patients under spinal cord stimulation are largely unknown.

%% file: sections/methods.tex
\section{METHODS} \label{sec:methods}

\subsection{Human Experiments}

\begin{figure}[ht]
\centering
\includegraphics[width=0.42\textwidth]{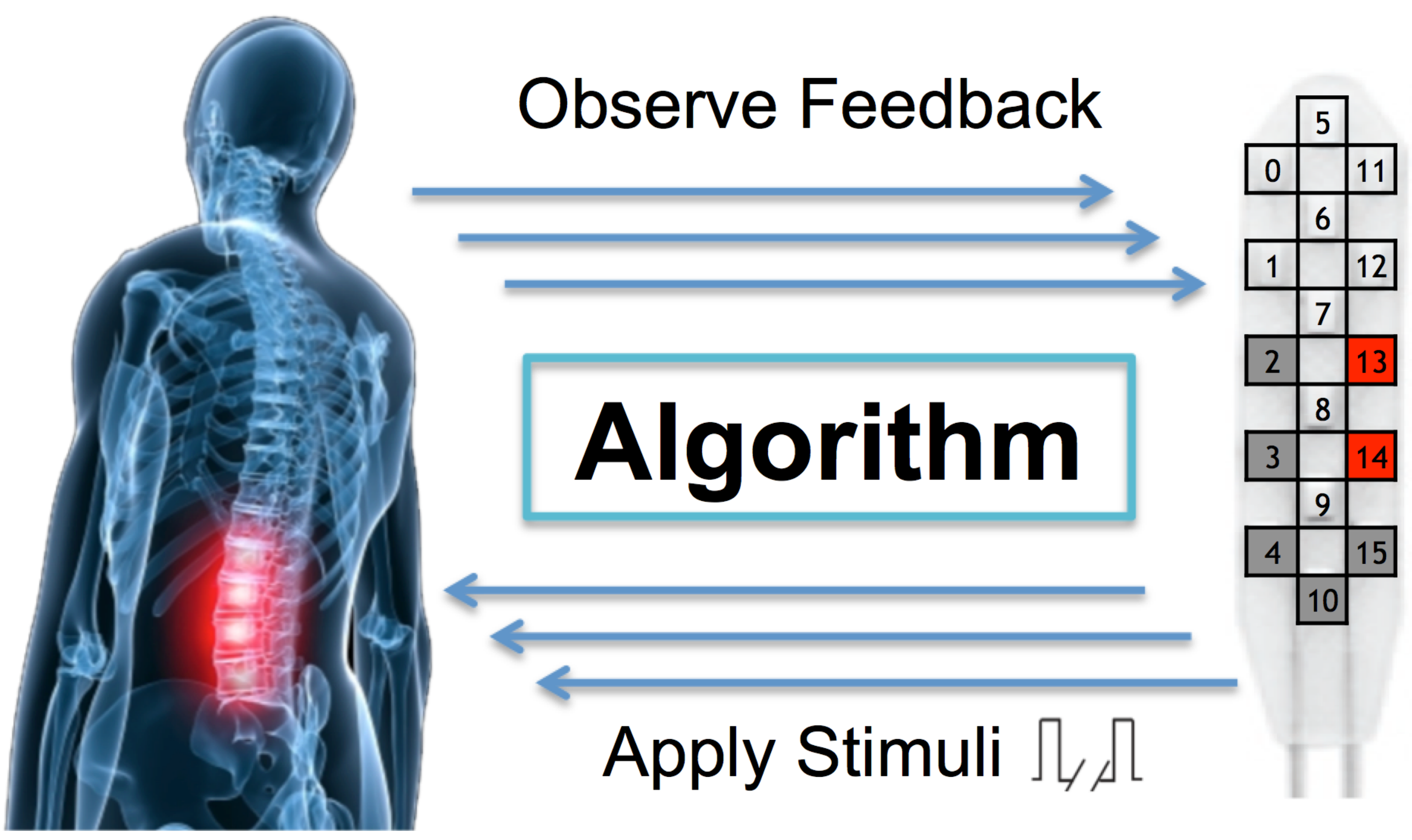}
\caption{Diagram of epidural stimulation process.}
\label{fig:exp}
\vskip -0.2 true in
\end{figure}

A conceptual illustration of the overall experimental process with a spinally stimulated SCI subject is shown in Fig. \ref{fig:exp}. The subject practices standing under spinal stimulation using a stand frame for assistance in achieving balance. An example of an electrode stimulation pattern is shown on the right of Fig.~\ref{fig:exp}. Each stimulus is a combination of active electrode selections (red and gray sites), the polarity of the actively selected electrodes (red as anodes and gray as cathodes), and the stimulation amplitude and frequency. Within each experiment, a different stimulus is chosen by active learning algorithms \cite{sui2015safe,sui2017correlational} and applied through the implanted electrode array and its controlling circuitry. Through out the whole experiment, a variety of different stimulation patterns have been tested. The standing quality under stimulation ranged from max-assisted standing to independent standing. Multi-channel EMG signals were recorded and quantitative scores for standing were provided by clinicians. A short video in the supplementary materials shows the standing performance under an effective stimulating pattern. 

The participant is under stable medical condition and has no painful musculoskeletal dysfunction that might interfere with the training. He has no motor response present in leg muscles during transcranial magnetic stimulation, indicating that there are no strongly active neural pathways connecting the cortex to lower limb muscles. No volitional control can be achieved during voluntary movement attempts in leg muscles as measured by EMG activity.

A total of 109 experimental trials were done with the same patient. Each trial lasted 5 minutes. Within each trial, one particular stimulating pattern was applied to the 16-channel electrode. The patterns were unchanged within each trial. For a fixed stimulation pattern, the stimulation frequency and amplitude were modulated synergistically in order to find the best values for effective weight-bearing standing. 

All EMG signals were sampled and recorded at 2000 Hz. Signals from right (R) and left (L) gluteus maximus (GL), medial hamstring (MH), vastus lateralis (VL), tibialis anterior (TA), medial gastrocnemius (MG) and soleus (SOL) were recorded by surface EMG electrodes. These 6 muscle groups are activated during standing and walking motion. 

The patient performed experimental training sessions for standing using a custom designed standing frame comprising horizontal bars anterior and lateral to the individual. These bars were used for upper extremity support and balance assistance as needed. If the knees or hips flexed beyond a safe standing posture, external assistance was provided at the knees to promote extension, and at the hips to promote hip extension and anterior tilt. Assistance was provided either manually by a trainer or by elastic bungee cords, which were attached between the two vertical bars of the standing apparatus. Mirrors were placed in front of the participant and laterally to him, in order to allow a better perception of the body position via visual feedback, conditioned on the lack of proprioceptive sensory feedback.

Stimulation began while the patient was seated. Then the participant initiated the sit to stand transition by positioning his feet shoulder width apart and shifting his weight forward to begin loading the legs. As shown in Fig. \ref{fig:exp}, the participant used the horizontal bars of the standing apparatus during the transition phase to balance and to partially pull himself into a standing position. Trainers positioned at the pelvis and knees manually assisted as needed during the sit to stand transition.

During sitting, negligible EMG activity of lower limb muscles was induced by epidural stimulation, showing that the weight-bearing related sensory information was needed to generate sufficient EMG patterns to effectively support full weight-bearing standing in spinally stimulated SCI.

Table \ref{table:score} describes how the clinicians quantified standing quality. Traditional measurements like center of pressure (COP) and center of mass (COM) cannot characterize the standing for paralyzed patients sufficiently. Typically, spinal cord injured patients do not stand and balance like normal subjects. Since there are no widely accepted quantitative measures for standing quality of paralyzed patients, we developed a discrete scoring system that ranges from $1$ to $10$. From scores $1$ to $5$, the standing is not independent but requires less and less assistance by bungees or trainers. With limited experimental resources, the max/mod/min level of assistance is the most robust quantification we could obtain from experienced assisting therapists. From scores $6$ to $10$ standing is overall independent and full-weight bearing. As the score increases, standing is more natural, stable, and durable. After every trial, a score on the overall standing quality was assigned. Both video and multi-channel EMG were recorded during the experiments. 

\begin{table}[ht]
\caption{The Scoring Criterions}
\vskip -0.15 true in
\label{table:score}
\begin{center}
\begin{small}
\begin{tabular}{|l|l|}
\hline
Score  & Descriptions \\ \hline
\hline
1-2 & Assisted by bungees or trainers (max)      \\ \hline
3-4 & Assisted by bungees or trainers (mod)      \\ \hline
5 & Assisted by bungees or trainers (min)      \\ \hline
6-7 & Hip: Not assisted, back arched \\ & Knee: Not assisted, loss of extension during shifting       \\ \hline
8-10  & Hip: Not assisted, back straight \\ & Knee: Not assisted, extended during shifting       \\ \hline
\end{tabular}
\end{small}
\end{center}
\vskip -0.15 true in
\end{table}

The research participants signed an informed consent for electrode implantation, stimulation, and physiological monitoring studies approved by the University of Louisville and the University of California, Los Angeles Institutional Review Boards. The individuals in this manuscript have also given written informed consent to publish these case details.

\subsection{Standing Model}

\begin{figure}[ht]
\centering
\includegraphics[width=0.21\textwidth]{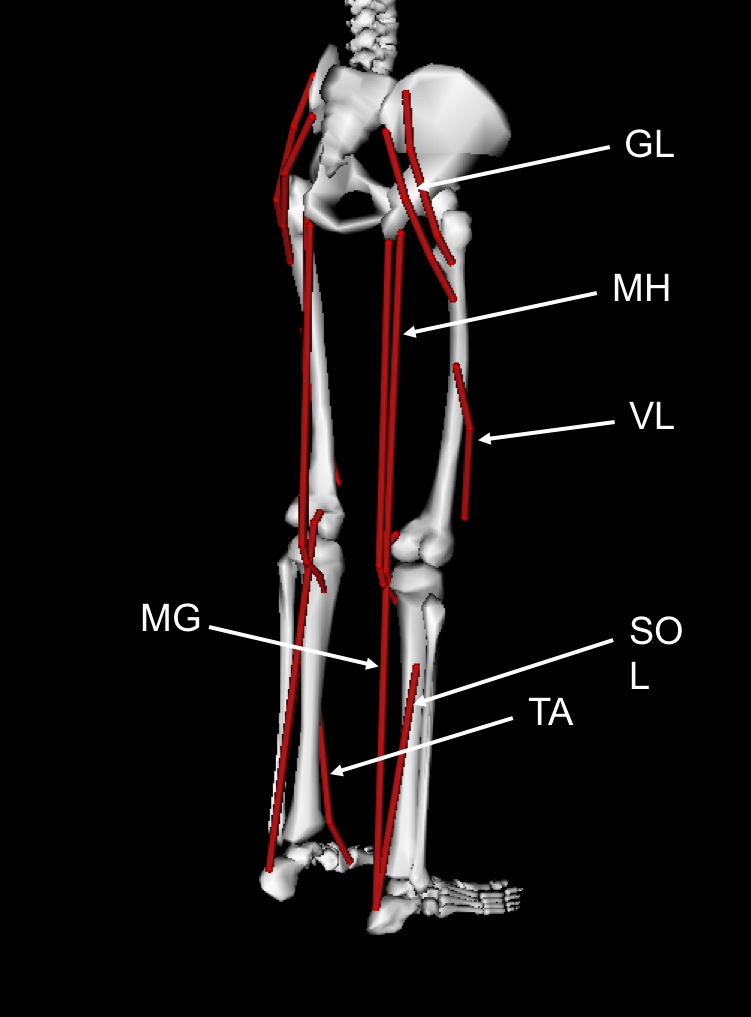}
\caption{Musculo-skeletal Model.}
\label{fig:model}
\vskip -0.12 true in
\end{figure}

Fig. \ref{fig:model} shows a simple musculoskeletal model of the legs and trunk used in this work (generated in OpenSim\cite{delp2007opensim}). It depicts the locations of uniarticular muscle tendon units (MTU) and the joints they actuate. The hip joint is extended by the gluteal muscles (GL) and flexed by the hip flexor muscles (HFL), while the knee joint is extended by the vastus lateralis (VL) and flexed by medial hamstring (MH). The tibialis anterior (TA) and the soleus (SOL) generate dorsiflexion and plantarflexion torques at the ankle, respectively. The Medial gastrocnemius (MG) is also considered. The choice of muscles is based on previous clinical experiments and the planar model proposed in \cite{geyer2010muscle}. 

For control of standing, this model may be redundant, as a subgroup of muscles \{GL, VL, SOL, TA\} can maintain stable posture. We  evaluate the redundancy of multi-channel EMG signals for predicting standing quality in Section \ref{sec:results}.

\subsection{EMG Processing}

\begin{figure*}[ht]
\vskip -0.1 true in
\centering
\includegraphics[width = 1\textwidth]{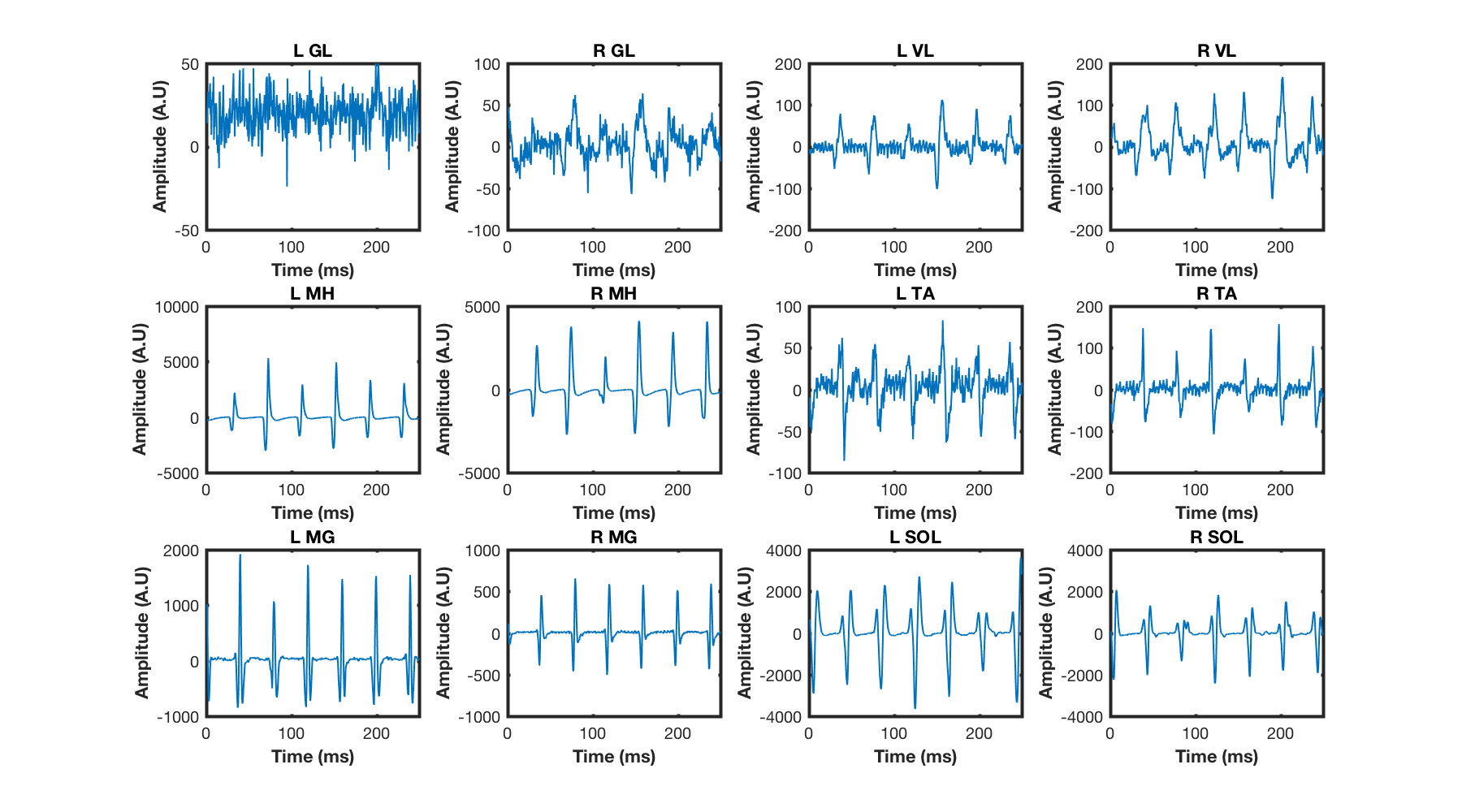}
%
\vskip -0.25 true in
\caption{12-channel EMG Signal. 'R GL' represents muscle GL on the right leg, etc. Amplitudes are not unified for better comparison of waveforms.}
\label{fig:signal}
\vskip -0.2 true in

\end{figure*}

\textbf{Feature Selection.} 
The 12-channel EMG signal of one experiment is shown in Fig. \ref{fig:signal} for a single experiment trial. Traditional methods such as time-domain and frequency-domain analyses have been widely utilized in EMG pattern recognition \cite{phinyomark2012feature}, and they have a good capability to track muscular changes. Other methods like Bayesian estimation \cite{sanger2007bayesian} and linear filtering also achieve good estimates of muscle forces. We first consider simple and robust linear models with one estimator per channel. For each EMG channel, we calculate the mean power within 50 seconds at the early stage of standing and use it as the only feature for that channel. These 12 features were extracted in each trial and used in LDA and linear regression models for simple and robust predictions.

\begin{figure}[]
\vskip -0.1 true in
\centering
\includegraphics[width=0.45\textwidth]{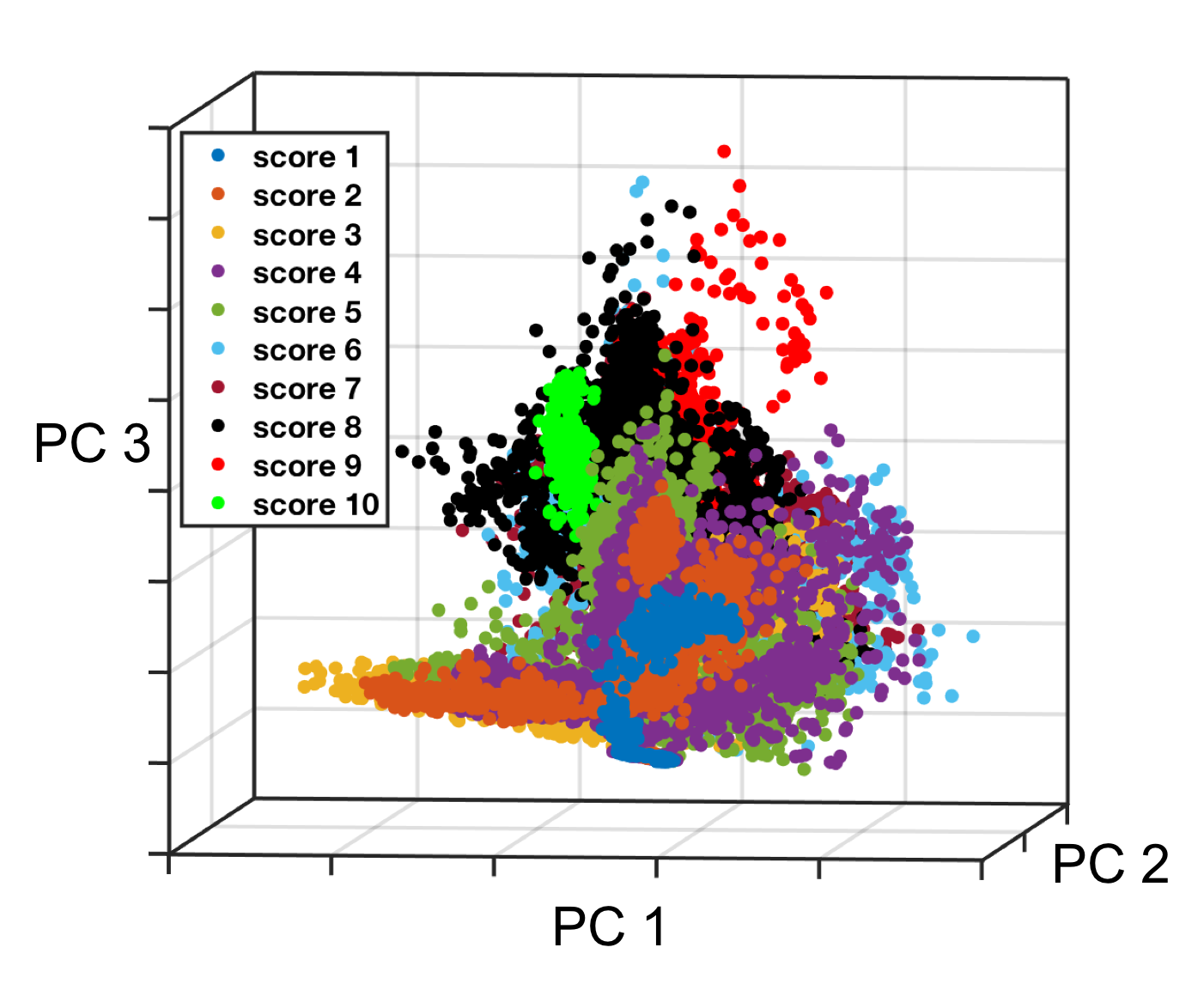}
\vskip -0.2 true in
\caption{First 3 principal components of EMG dataset}
\label{fig:scatter}
\vskip -0.25 true in
\end{figure}

For multi-class SVM feature selection, we drew inspiration from previous work that implementing machine learning techniques to predict forces applied at joints using EMG signals for exoskeleton control \cite{khokhar2010surface}. A $4^{th}$ order Auto-Regressive(AR) model was fit to a 250 ms window of each EMG channel and the four coefficients (excluding the bias) were extracted as features. Thus, for 12-channels, a total of 48 features were extracted per observation. By performing 10-fold cross validation on the optimum number of principal components we reduced the training set to the top $19$ dimensions which capture $98\%$ of the variance. Fig. \ref{fig:scatter}. shows the standing scores plotted against the first three principal components of the SVM training set. Even in 3-dimensions we see a high-degree of separability.

\textbf{Model Selection.}
As shown in Table \ref{table:score}, the data can be coarsely fit into 2 groups: good performances (not assisted, with score $>$ 5) and bad performances (assisted, with score $\leq $ 5). We apply linear discriminant analysis (LDA) on the 2-class training data and predict whether a new group of EMG signals represents good or bad standing performance. We further train the kernel-SVM model for better accuracy. The SVM is trained to directly predict the fine-grained standing quality score by translating the problem to a multi-class classification task with 10 classes (scores $1-10$). Each standing score corresponds to one class. A radial basis function with a scaling factor $\gamma = 0.79$ is used for the SVM kernel and a box constraint level of $C = 11$ was used to control the number of support vectors. 

To show the robustness of EMG signals, we also estimate standing quality by directly applying linear regression on the clinician scores v.s 12-dimension power features.

%% file: sections/results.tex
\section{RESULTS} \label{sec:results}

\subsection{Estimating Standing Qualities}

The 12 surface EMG channels represent 6 muscle groups (GL, VL, MH, TA, MG, SOL) for both legs. An EMG waveform recorded during a high performing stance is shown in Fig. \ref{fig:signal}. 'R GL' represents right leg muscle GL, etc. The majority of muscles have strong and stationary EMG signals.

We first apply the LDA model with 12-dimension power features as input. The classification of good or bad performances yields an accuracy of $89.91\%$, which is quite high given the usage of a simple LDA model on a limited number of features. This classifier is already sufficient to be used in practice for fast and robust assessment of standing quality.

\begin{figure}[ht]
\vskip -0.1 true in
\centering
\includegraphics[width=0.45\textwidth]{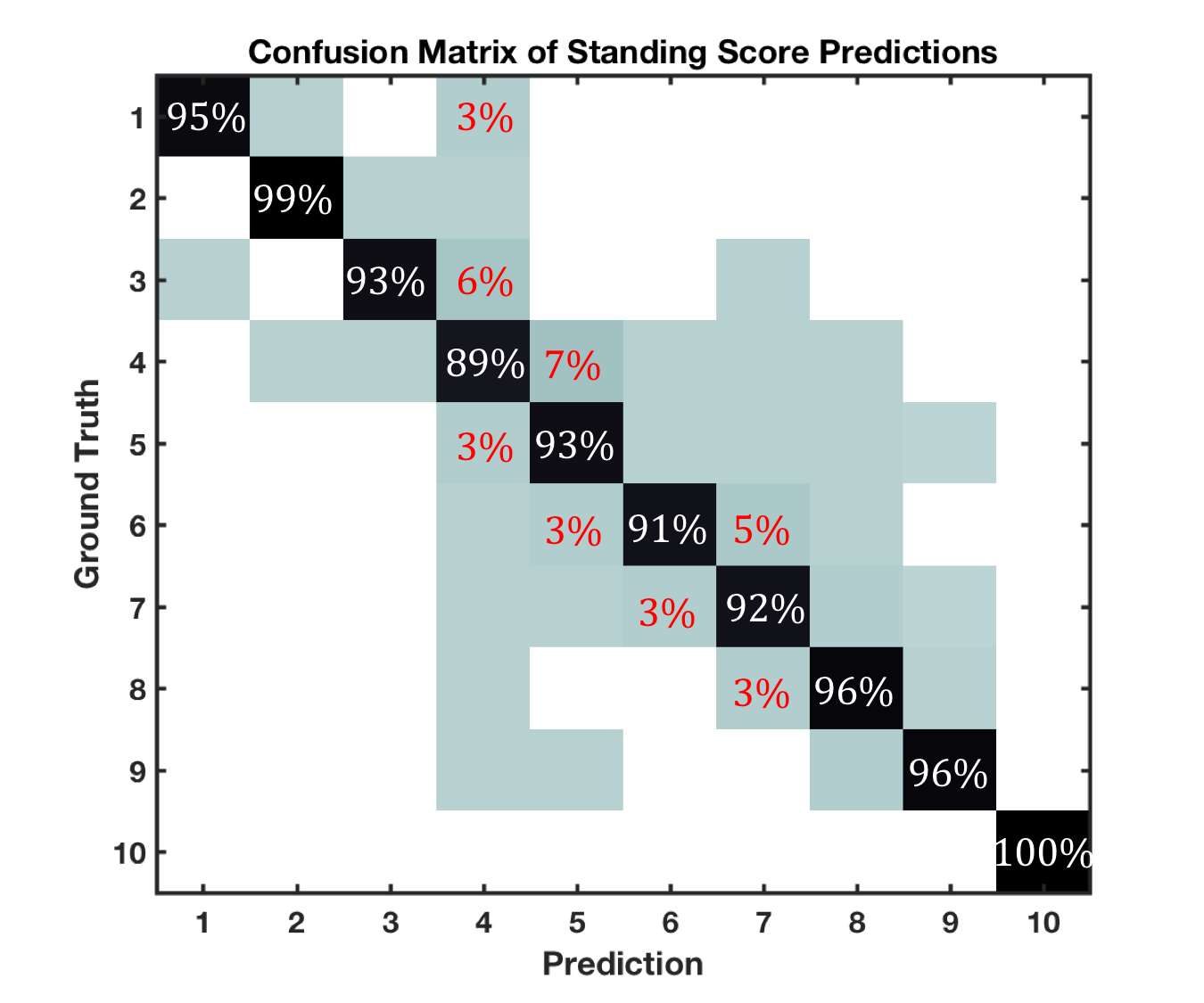}
\vskip -0.15 true in
\caption{Confusion matrix of predictions made with SVM}
\label{fig:confusion}
\vskip -0.1 true in
\end{figure}

The kernel-SVM model yields $93.9\%$ classification accuracy on the 10-class discrimination task upon 10-fold cross validation, which confirms our belief that EMG signals are accurate predictors of bipedal standing. Moreover, we see that using a more sophisticated model enables us to achieve higher classification accuracy then the linear model even with more classes. From the confusion plot in Fig. \ref{fig:confusion}, we see that most predictions lie within the range of the super diagonals indicating that it is highly unlikely for the SVM to mis-predict a score by a difference greater than 1. White percentages indicate the correct predictions and red percentages denote the misprediction rates. Each row sums to 1. The slots with misprediction rates less than $3\%$ were omitted for succinctness. A standing score of 4 is the most often mis-predicted class due to it's similarity with score 5 which can be attributed to the fact that these scores lie on the boundary between the mod and min level of assistance.

\begin{figure}[ht]
\vskip -0.1 true in
\centering
\includegraphics[width=0.4\textwidth]{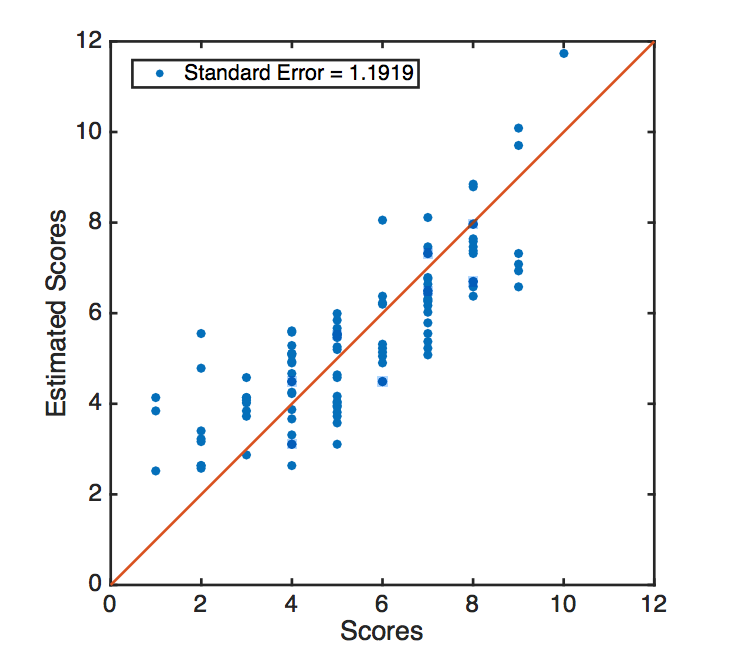}
\vskip -0.05 true in
\caption{Regression on the Scores with 6 pairs (12-channel) EMG.} 
\label{fig:regression}
\vskip -0.1 true in
\end{figure}

To estimate the score for each experiment from EMG features, linear regression is applied with 12-channel power features as inputs. In Fig. \ref{fig:regression}, the $x$-axis represents true scores and the $y$-axis plots the linear regression estimate. The red line represents a perfect match, $y=x$. Each dot represents one experiment's true and estimated score. The dots should scatter near  the red line if the estimator performs well. Within the 109 experiments, $57.8\%$ of the estimates are within $\pm 1$ of the true score, $93.6\%$ of the estimates are within the region of true score $\pm 2$, and $98.2\%$ of the estimates are within the region of true score $\pm 3$. The standard deviation of estimation errors is $1.19$, which is modest in the $1$-$10$ scoring range. 

\subsection{Reducing EMG Channels}

Although more EMG channels provide better estimation, in practice we may not have the access to as many channels. Also, fewer channels improve experimental  efficiency and cost. We investigate the possibility of reducing the number of EMG channels while maintaining a high accuracy rate. 

To choose optimal $k$-muscle sub-groups based on the $6$ muscle group recordings, we evaluate the classification/regression performance of all ${6 \choose k}$ muscle combinations. The optimal combination for each $k$ is shown in Table \ref{table:order}. The single best muscle for prediction is soleus (SOL).

\begin{table}[ht]
\vskip -0.05 true in
\caption{The Optimal Reduced Order EMG Channels}
\vskip -0.1 true in
\label{table:order}
\begin{center}
\begin{small}
\begin{tabular}{|l|l|}
\hline
Num. of Pairs & Optimal Combinations of EMG Channels \\ \hline
\hline
6 & GL, VL, MH, TA, MG, SOL      \\ \hline
5 & VL, MH, TA, MG, SOL      \\ \hline
4 & VL, TA, MG, SOL      \\ \hline
3 & VL, MH, SOL       \\ \hline
2 & VL, SOL       \\ \hline
1 & SOL      \\ \hline
\end{tabular}
\end{small}
\end{center}
\vskip -0.2 true in
\end{table}

Notice, this reduction process is different from principal component analysis (PCA) which reduces the feature space by picking the top independent components. Our approach aims at achieving good classification/regression by using fewer number of EMG channels. The chosen EMG channels may not be independent with each other.

Table \ref{table:accuracy} shows the optimal classification results with different number of muscle groups for both 2-class LDA and 10-class SVM classification. For both models, the accuracy slowly decreases as the number of chosen muscle groups ($k$) decreases from $k=6$ to $k=2$. A high accuracy of $87.16\%$ (for LDA), and $89.5\%$ (for SVM) is maintained even at $k = 2$. The muscle groups vastus lateralis (VL) and soleus (SOL) are the optimal combination for $k = 2$. One of them (SOL) is the ankle flexor and the other (VL) is a knee extensor (see Fig. \ref{fig:model}). The accuracies drop significantly from $k = 2$ to $k = 1$. This makes sense since at least 2 actuators are needed to control the 2 degrees of freedom. 

\begin{table}[ht]
\caption{The Accuracies with Reducing Channels}
\label{table:accuracy}
\vskip -0.5 true in
\begin{center}
\begin{small}
\begin{tabular}{|l|l|l|}
\hline
Channels v.s.Accuracies & LDA(2-class) & SVM(10-class)  \\ \hline
\hline
GL, VL, MH, TA, MG, SOL & 89.91\% & 93.9\%      \\ \hline
VL, MH, TA, MG, SOL & 88.91\% & 93.6\%      \\ \hline
VL, TA, MG, SOL & 88.07\% & 93.0\%      \\ \hline
VL, MH, SOL & 87.16\% & 92.7\%       \\ \hline
VL, SOL  & 87.16\%  & 89.5\%     \\ \hline
SOL  & 80.73\% & 63.5\%      \\ \hline
\end{tabular}
\end{small}
\end{center}
\vskip -0.2 true in
\end{table}




%% file: sections/conclusions.tex
\section{DISCUSSION AND CONCLUSION} \label{sec:conclusions}

\subsection{Predictions}
This paper showed that multi-channel EMG recordings can provide accurate, fast, and robust estimation for the quality of bipedal standing under spinal stimulation. 
The prediction performance under reduced numbers of EMG channels was also experimentally evaluated, which confirms our hypothesis that 12-channel EMG signals are  redundant for predicting standing quality. We also found the optimal combinations of reduced muscle channels, and further demonstrate that performance can be maintained at a high level even with few channels. This fact challenges our initial assumption of multiple muscle group coordination.  However, we believe that better estimation of standing quality requires recording from a larger number of muscle groups. 

There are multiple ways to improve the accuracy of the predictions. We have demonstrated that using more elaborate features with a SVM allows realizes higher prediction accuracy for the multi-class problem. Including more EMG signal features and using finer picked/tuned models is one venue for improvement. As mentioned in Section \ref{sec:introduction}, the activity of major muscle groups during spinal stimulated standing can be very different from the activity of natural standing. Stimulated standing EMG  contains an early response strongly modulated by the electrical stimulation, and a late response which is more like the EMG patterns of healthy subjects. Separating these two stages should also improve prediction accuracy. Under a constraint on the number of EMG channels, asymmetric placement of EMG sensors on left and right legs could also improve prediction accuracy, assuming the spinal stimulation affects both sides equally. We could also adding more physical sensors, such as accelerometer.

In general, our estimators can provide reliable scores on the quality of patient standing when experienced clinicians  are not available during experiments.

\subsection{Sensor Placement Efficiency}
We investigated optimal EMG channel combinations under a constraint on the number of channels. What if more EMG are available? Previous research \cite{gartman2008selection} suggested that a set of $8 \times 2$ muscles supports $42\%$ of standing postures. Coactivation of an extra $4 \times 2$ muscles increased the percentage of feasible postures to $71\%$. We can sample from this larger space, and it may reduce to a better group of 12 channels. 

\subsection{Combining with Exoskeleton}
Many SCI patients could benefit from rehabilitation robotic systems to provide functional gait therapies or assistance in standing and moving. Current assistive standing systems rarely incorporate feedback other than direct force measurements from users. Automated estimates of bipedal standing quality using EMG signals could provide useful signals to assistive systems for better standing control.